\pdfoutput=1

\documentclass[11pt]{article}
\usepackage{kotex} 
\usepackage{makecell} 
\usepackage[final]{coling}%

\usepackage{times}
\usepackage{latexsym}

\usepackage[T1]{fontenc}

\usepackage[utf8]{inputenc}

\usepackage{microtype}

\usepackage{inconsolata}

\usepackage{graphicx}

\title{Can Large Language Models Predict Antimicrobial Resistance Gene?}

\author{Hyunwoo Yoo \\
  Drexel University \\
  \texttt{hty23@drexel.edu}}

\begin{document}
\maketitle
\begin{abstract}

This study demonstrates that generative large language models can be utilized in a more flexible manner for DNA sequence analysis and classification tasks compared to traditional transformer encoder-based models. While recent encoder-based models such as DNABERT and Nucleotide Transformer have shown significant performance in DNA sequence classification, transformer decoder-based generative models have not yet been extensively explored in this field. This study evaluates how effectively generative Large Language Models handle DNA sequences with various labels and analyzes performance changes when additional textual information is provided. Experiments were conducted on antimicrobial resistance genes, and the results show that generative Large Language Models can offer comparable or potentially better predictions, demonstrating flexibility and accuracy when incorporating both sequence and textual information. The code and data used in this work are available at the following GitHub repository: https://github.com/biocomgit/llm4dna.

\end{abstract}

\section{Introduction}

Language Models have shown notable performance in various tasks in Natural Language Processing and have recently been applied to bioinformatics tasks such as DNA sequence analysis. Encoder-based transformer models trained on nucleotide sequences, such as DNABERT \cite{Ji2021DNABERT, zhou2023dnabert2} and Nucleotide Transformer \cite{dallatorre2023nucleotide}, have demonstrated excellent performance in DNA sequence classification and are widely used for various gene sequence analyses. Additionally, encoder-based models trained on amino acid sequences have also shown good performance in protein sequence classification \cite{Brandes2022ProteinBERT}. However, generative Large Language Models, such as GPT based models\cite{brown2020gpt3}, have not yet been actively utilized for DNA analysis. While models like BioGPT \cite{luo2022biogpt}, which are primarily trained on medical and pharmaceutical text, have emerged, models focused specifically on DNA analysis are still relatively rare.

Generative large language Models possess the flexibility to make predictions through appropriate prompts, even when incorporating additional textual information, which can significantly enhance performance. This ability to directly utilize supplementary text in the training process sets them apart from encoder-based models, which often struggle with handling complex textual data. Furthermore, generative models can also adapt to situations where the same DNA sequence may have multiple labels, offering a level of versatility not typically seen in models relying on fixed labels.

This study primarily investigates how the performance of generative large language models improves when supplementary textual information is provided alongside DNA sequences. Additionally, it explores how these models handle cases where a single DNA sequence may be associated with different labels. Through this, the study aims to demonstrate the potential applications of generative language models in DNA analysis.

\section{Related Works}

This study is connected to existing research on various DNA sequence analyses and the classification of antibiotic resistance genes. 

\textbf{AMR++} introduces an updated database of antimicrobial resistance determinants along with a classification pipeline, which classifies genes based on large-scale sequencing data \cite{Bonin2023MEGAResAMR}. This study designs experimental datasets for DNA sequence analysis using this data.

\textbf{AMR-meta} presents a method for analyzing high-speed single-read metagenomic data using k-mers and meta-features \cite{marini2022amrmeta}, contributing to the classification of antibiotic resistance genes. This study builds on this by conducting analyses using various types of metadata.

\textbf{Meta-MARC} proposes a method for detecting antibiotic resistance gene sequences using hierarchical hidden Markov models, which serves as an important reference for DNA sequence analysis dealing with diverse sequences \cite{lakin2019hierarchical}.

\textbf{DeepARG} introduces a deep learning method for predicting antibiotic resistance genes, effectively classifying these genes using metagenomic data \cite{arangoargoty2018deeparg}. This study extends such foundational deep learning-based DNA sequence analysis methods to explore how generative language models can be applied to gene analysis.

\textbf{Blastn} is a tool that compares the similarity between DNA sequences, performing local sequence alignments between query sequences and target sequences stored in databases. Since the database contains sequences already annotated with gene functions, Blastn can predict genetic functions by finding sequences similar to the query sequence. It divides the query sequence into small fragments (words), finds matching sequences, extends them, and evaluates the similarity by providing results with an E-value. It is widely used in bioinformatics for gene function prediction, evolutionary relationship analysis, and sequence variation detection \cite{lobo2008blast}.

\textbf{DNABERT} is one of the first models to apply BERT to DNA sequence analysis. Based on the BERT architecture, which is primarily used in Natural Language Processing, DNABERT tokenizes DNA sequences and learns the meaning of each nucleotide sequence \cite{Ji2021DNABERT}. The main technique of DNABERT1 is the use of k-mer tokenization, where DNA nucleotide sequences are divided into 3-mers or 6-mers to process them like words in natural language. This helps better understand the relationships between sequences and extract patterns within genes. DNABERT maintains BERT’s pre-training and fine-tuning techniques while performing pre-training on large-scale DNA sequence data, resulting in high performance in tasks such as promoter prediction, splicing site detection, and DNA sequence classification. The self-attention mechanism of the transformer model allows DNABERT to effectively learn long-term dependencies within DNA sequences, performing similarly or slightly better than traditional rule-based models or simple sequence alignment algorithms.

\textbf{DNABERT2}, while maintaining the BERT-based transformer architecture, includes more biological data in its pre-training, enabling the model to learn complex genetic interactions and fine DNA patterns \cite{zhou2023dnabert2}. DNABERT2 overcomes the inefficiencies of traditional k-mer tokenization by introducing the Byte Pair Encoding method for DNA sequence analysis. While the k-mer method used fixed-length sequence fragments as tokens, leading to inefficiencies in processing, DNABERT2 implements BPE to merge frequently occurring sequences, enabling more efficient non-overlapping tokenization. This allows the model to overcome input length limitations, reduce time and memory usage, and improve performance.

\textbf{Nucleotide Transformer} is a transformer-based model pre-trained on various human and non-human genomic data, showing excellent performance in predicting molecular phenotypes from DNA sequences. The model is trained on 3,202 human genomes and 850 genomes from various species, generating context-specific representations \cite{dallatorre2023nucleotide}. These representations allow for high-accuracy predictions even with small amounts of data, matching or surpassing existing specialized methods in 11 of 18 prediction tasks. After fine-tuning, the performance improved in 15 tasks. 
Nucleotide Transformer focuses on learning important genetic elements, such as enhancers, that regulate gene expression, and it has shown the ability to identify such elements without supervised learning. Additionally, it has been demonstrated that utilizing the model’s representations can improve the prioritization of functional genetic variants.

\begin{table}
  \centering
  \begin{tabular}{lc}
    \hline
    \textbf{Model} & \textbf{Unclassified Rate} \\
    \hline
    \makecell{LLama3.1 8B-4bit \\ (Base Model)}     & 97\%           \\
    \makecell{LLama3.1 8B-4bit \\ (Blastn)}  & 73\%            \\
    \makecell{LLama3.1 8B-4bit \\ (Finetuning)} & 0\%           \\
    \makecell{Claude3.5sonet  \\ (Base Model)}   & 39\%            \\
    \makecell{Claude3.5sonet  \\ (Blastn)}    & 11\%           \\
    \makecell{Chatgpt4o-mini  \\ (Base Model)}   & 100\%            \\
    \makecell{Chatgpt4o-mini  \\ (Blastn)}    & 14\%           \\
    \makecell{Chatgpt4o-mini  \\ (Finetuning)}   & 0\%            \\
    
    \hline
  \end{tabular}
  \caption{Model unclassified rates with long names displayed in two lines.}
  \label{tab:model-unclassified-rate}
\end{table}

\begin{table*}
  \centering
  \begin{tabular}{lcccc}
    \hline
    \textbf{Model}           & \textbf{Accuracy} & \textbf{Precision} & \textbf{Recall} & \textbf{F1 Score} \\
    \hline
    
    LLama3.1 8B-4bit       & 0.0037           & 0.0011           & 0.0002           & 0.0003  \\
    LLama3.1 8B-4bit + Blastn & 0.0744           & 0.0530           & 0.0129           & 0.0207  \\
    LLama3.1 8B-4bit + Finetuing & 0.5521           & 0.4760           & 0.5521           & 0.5080  \\
    Claude3.5sonet       & 0.1488           & 0.1770           & 0.0966           & 0.0735  \\
    Claude3.5sonet + Blastn & 0.8042           & 0.6287           & 0.5421           & 0.5794  \\
    Chatgpt4o-mini       & 0.00           & 0.00           & 0.00           & 0.00           \\
    Chatgpt4o-mini + Blastn       & 0.7804           & 0.9090           & 0.7804     & 0.8398  \\
    Chatgpt4o-mini + Finetuning & 0.9318           & 0.9337           & 0.9318     & 0.9319    \\
    
    \hline
  \end{tabular}
  \caption{\label{model-performance}
    Performance metrics for various large language models.
  }
  \label{tab:model-performance}
\end{table*}

\section{Methods}

\subsection{Data Collection}

The data used in this study consists of antibiotic resistance gene data collected from the MEGARes \cite{Doster2020} and CARD databases \cite{Jia2017CARD}. The different labels from MEGARes and CARD were integrated using the Antibiotic Resistance Ontology from the European Bioinformatics Institute, as described in previous research methods \cite{yoo2024predicting} \cite{cook2016european}. These databases contain DNA sequences of various antibiotic resistance genes, and each sequence is classified with one or more labels. Additionally, the BLASTn algorithm was used to search for antibiotic resistance sequences similar to each DNA sequence, and the top 5 search results were selected based on the e-value criterion.

\subsection{Preprocessing}
The collected DNA sequences underwent preprocessing for analysis. First, all DNA sequences were converted to uppercase, and invalid sequences were removed from the BLASTn results. The final dataset was constructed by including only antibiotic resistance gene sequences.

\subsection{Fine-tuning}

In this study, we applied the Low-Rank Adaptation (LoRa) technique to fine-tune the LLaMA model and the ChatGPT4-mini model for DNA sequence classification tasks. LoRa is an efficient method that converts only a portion of the parameters of large models into low-rank matrices for additional training, thereby reducing memory and computational costs while maintaining model performance \cite{hu2021lora}. Additional experiments were conducted using the Claude 3.5 sonet API, and the performance of various generative language models was compared and analyzed.

\section{Experiments}









\subsection{Setup}

In this study, we used the 8B version of the LLaMA 3.1 model\cite{Meta2024Llama31} with 4-bit quantization from the unsloth model \cite{han2023unsloth} as the foundation model for our experiments. Additionally, the LLaMA 405B model provided through the Amazon Bedrock API was utilized. Separate experiments were conducted with ChatGPT4-mini and the Claude 3.5 sonet API to evaluate the performance of these diverse language models.

\subsection{Metric}

The performance of the model was evaluated based on the accurate classification of antibiotic resistance genes in the DNA sequences. For fine-tuned models, this criterion was used since they mostly returned the correct class labels in the output. However, for non-fine-tuned models, they rarely returned just the label names. Therefore, we implemented a model that maps class labels from rather lengthy and verbose explanations, and using this model, we extracted the class labels again. The classification performance of the large language models was then evaluated based on these labels. In addition to classification accuracy, the models were also evaluated using precision, recall, and F1 score.

\section{Results}

The experimental results showed that generative large language models performed just as well as traditional encoder-based models in handling multiple labels for DNA sequences. Furthermore, when additional gene information from the Blastn DB search results was provided, performance improved even without additional training on this data. As seen in Table \ref{tab:model-unclassified-rate}, the Unclassified Rate decreased across all models. For the LLaMA 3.1 8B-4bit quantized model, the rate dropped from 97\% to 73\% when using Blastn. For Claude 3.5 sonet, it decreased from 39\% to 11\%. ChatGPT 4-mini showed a sharp improvement, going from classifying nothing to only leaving 14\% unclassified. When fine-tuning was applied, both the LLaMA 3.1 8B 4bit quantized model and ChatGPT 4-mini reduced their unclassified rates to 0\%.

As shown in Table \ref{tab:model-performance}, overall classification performance also improved when using additional Blastn information or fine-tuning. For example, the accuracy of the LLaMA 3.1 8B-4bit model increased from 0.0037 to 0.0744 when using Blastn, and to 0.5521 with fine-tuning. The F1 Score followed a similar trend. Claude 3.5 sonet and ChatGPT 4-mini also showed improvements in accuracy, precision, recall, and F1 score when using the additional Blastn information. With fine-tuning, the accuracy of the ChatGPT 4-mini model rose to 0.9318, comparable to language models trained solely on DNA data. Without additional information or fine-tuning, all metrics for the ChatGPT 4-mini model were zero, indicating that the model made no predictions. The model's responses suggested that it could not make a judgment due to insufficient evidence.

Table \ref{tab:model-accuracy} shows the classification accuracy on test data based on the CARD labels without any additional information or training. The test was conducted using models fine-tuned on the integrated MEGARes-based label data, and the accuracy was 0.2307 for the LLaMA 3.1 8B-4bit model and 0.5037 for the ChatGPT 4-mini model. This demonstrates that large language models can still handle different labeling environments to some extent.

Large language models not only flexibly incorporate additional textual information but also improve their performance with this data. Moreover, they can handle previously unseen labels to some extent, and fine-tuning further enhances their performance. This shows that generative language models are advantageous in solving more complex problems, such as classifying DNA AMR drug classes, by integrating diverse information.

\section{Conclusion}

This study confirmed that generative large language models can be applied more flexibly to DNA sequence analysis and classification tasks compared to traditional encoder-based models. In particular, we found that model performance improved when additional textual information was provided, offering important implications for future DNA analysis and antibiotic resistance gene classification tasks. Additionally, experiments showed that generative models are more flexible and accurate in handling cases where different labels exist for the same sequence, even if these labels differ from those in the training data.

\section{Discussion}
This study highlighted the flexibility of generative large language models and their ability to utilize additional textual information, showing that they can play an important role in DNA sequence analysis. Generative models have the potential to handle unseen labels effectively and integrate supplementary information to solve more complex problems. However, this study used a limited dataset, and future research should expand the scope of analysis by including a wider range of DNA sequences and antibiotic resistance gene data.

Extracting specific class labels from long texts was not a straightforward task. Initially, the final prediction was based on the antibiotic name most frequently mentioned in the model's output. In cases where multiple antibiotics were predicted equally, they were excluded from evaluation, but this method was not very accurate. To address this, a model was implemented to extract class labels from lengthy texts for evaluation. However, the evaluation may vary depending on the accuracy of this extraction model.

\begin{table}[h]
  \centering
  \begin{tabular}{lc}
    \hline
    \textbf{Model} & \textbf{Accuracy} \\
    \hline
        \makecell {LLama 3.1 8B-4bit \\ (Fine-tuning)}     & 0.2307    \\
        \makecell{Chatgpt 4o-mini \\ (Fine-tuning)}  & 0.5037   \\    
    \hline
  \end{tabular}
  \caption{Model accuracy results with different label dataset.}
  \label{tab:model-accuracy}
\end{table}

\bibliography{custom}

\begin{thebibliography}{18}
\providecommand{\natexlab}[1]{#1}

\bibitem[{AI(2024)}]{Meta2024Llama31}
Meta AI. 2024.
\newblock Introducing llama 3.1: Our most capable models to date.
\newblock \url{https://ai.meta.com/blog/meta-llama-3-1/}.
\newblock Accessed: 2024-09-06.

\bibitem[{Arango-Argoty et~al.(2018)Arango-Argoty, Garner, Pruden, Heath, Vikesland, and Zhang}]{arangoargoty2018deeparg}
Gustavo Arango-Argoty, Emily Garner, Amy Pruden, Lenwood~S. Heath, Peter Vikesland, and Liqing Zhang. 2018.
\newblock \href {https://doi.org/10.1186/s40168-018-0401-z} {Deeparg: A deep learning approach for predicting antibiotic resistance genes from metagenomic data}.
\newblock \emph{Microbiome}, 6(1):23.

\bibitem[{Bonin et~al.(2023)Bonin, Doster, Worley, Pinnell, Bravo, Ferm, Marini, Prosperi, Noyes, Morley, and Boucher}]{Bonin2023MEGAResAMR}
Nathalie Bonin, Enrique Doster, Hannah Worley, Lee~J Pinnell, Jonathan~E Bravo, Peter Ferm, Simone Marini, Mattia Prosperi, Noelle Noyes, Paul~S Morley, and Christina Boucher. 2023.
\newblock \href {https://doi.org/10.1093/nar/gkac1047} {Megares and amr++, v3.0: an updated comprehensive database of antimicrobial resistance determinants and an improved software pipeline for classification using high-throughput sequencing}.
\newblock \emph{Nucleic Acids Research}, 51(D1):D744--D752.

\bibitem[{Brandes et~al.(2022)Brandes, Ofer, Peleg, Rappoport, and Linial}]{Brandes2022ProteinBERT}
Nadav Brandes, Dan Ofer, Yam Peleg, Nadav Rappoport, and Michal Linial. 2022.
\newblock \href {https://doi.org/10.1093/bioinformatics/btac020} {Proteinbert: a universal deep-learning model of protein sequence and function}.
\newblock \emph{Bioinformatics}, 38(8):2102--2110.

\bibitem[{Brown et~al.(2020)Brown, Mann, Ryder, Subbiah, Kaplan, Dhariwal, Neelakantan, Shyam, Sastry, Askell, Agarwal, Herbert-Voss, Krueger, Henighan, Child, Ramesh, Ziegler, Wu, Winter, Hesse, Chen, Sigler, Litwin, Gray, Chess, Clark, Berner, McCandlish, Radford, Sutskever, and Amodei}]{brown2020gpt3}
Tom~B. Brown, Benjamin Mann, Nick Ryder, Melanie Subbiah, Jared Kaplan, Prafulla Dhariwal, Arvind Neelakantan, Pranav Shyam, Girish Sastry, Amanda Askell, Sandhini Agarwal, Ariel Herbert-Voss, Gretchen Krueger, Tom Henighan, Rewon Child, Aditya Ramesh, Daniel~M. Ziegler, Jeffrey Wu, Clemens Winter, Christopher Hesse, Mark Chen, Eric Sigler, Mateusz Litwin, Scott Gray, Benjamin Chess, Jack Clark, Christopher Berner, Sam McCandlish, Alec Radford, Ilya Sutskever, and Dario Amodei. 2020.
\newblock \href {https://doi.org/10.48550/arXiv.2005.14165} {Language models are few-shot learners}.
\newblock \emph{arXiv preprint arXiv:2005.14165}.

\bibitem[{Cook et~al.(2016)Cook, Bergman, Finn, Cochrane, Birney, and Apweiler}]{cook2016european}
Charles~E. Cook, Mary~Todd Bergman, Robert~D. Finn, Guy Cochrane, Ewan Birney, and Rolf Apweiler. 2016.
\newblock \href {https://doi.org/10.1093/nar/gkv1352} {The european bioinformatics institute in 2016: Data growth and integration}.
\newblock \emph{Nucleic Acids Research}, 44(D1):D20--D26.

\bibitem[{Dalla-Torre et~al.(2023)Dalla-Torre, Gonzalez, Mendoza-Revilla, Carranza, Grzywaczewski, Oteri, Dallago et~al.}]{dallatorre2023nucleotide}
Hugo Dalla-Torre, Liam Gonzalez, Javier Mendoza-Revilla, Nicolas~Lopez Carranza, Adam~Henryk Grzywaczewski, Francesco Oteri, Christian Dallago, et~al. 2023.
\newblock \href {https://doi.org/10.1101/2023.01.11.523679} {The nucleotide transformer: Building and evaluating robust foundation models for human genomics}.
\newblock \emph{Genomics}.

\bibitem[{Doster et~al.(2020)Doster, Lakin, Dean, Wolfe, Young, Boucher, Belk, Noyes, and Morley}]{Doster2020}
Enrique Doster, Steven~M Lakin, Christopher~J Dean, Cory Wolfe, Jared~G Young, Christina Boucher, Keith~E Belk, Noelle~R Noyes, and Paul~S Morley. 2020.
\newblock \href {https://doi.org/10.1093/nar/gkz1010} {Megares 2.0: a database for classification of antimicrobial drug, biocide and metal resistance determinants in metagenomic sequence data}.
\newblock \emph{Nucleic Acids Research}, 48(D1):D561--D569.

\bibitem[{Han and Han(2023)}]{han2023unsloth}
Daniel Han and Michael Han. 2023.
\newblock {unsloth}.
\newblock \url{https://github.com/unslothai/unsloth}.

\bibitem[{Hu et~al.(2021)Hu, Shen, Wallis, Allen-Zhu, Li, Wang, Wang, and Chen}]{hu2021lora}
Edward~J. Hu, Yelong Shen, Phillip Wallis, Zeyuan Allen-Zhu, Yuanzhi Li, Shean Wang, Lu~Wang, and Weizhu Chen. 2021.
\newblock \href {https://arxiv.org/abs/2106.09685v2} {Lora: Low-rank adaptation of large language models}.
\newblock \emph{arXiv}.
\newblock ArXiv:2106.09685v2.

\bibitem[{Ji et~al.(2021)Ji, Zhou, Liu, and Davuluri}]{Ji2021DNABERT}
Yanrong Ji, Zhihan Zhou, Han Liu, and Ramana~V Davuluri. 2021.
\newblock \href {https://doi.org/10.1093/bioinformatics/btab083} {Dnabert: pre-trained bidirectional encoder representations from transformers model for dna-language in genome}.
\newblock \emph{Bioinformatics}, 37(15):2112--2120.

\bibitem[{Jia et~al.(2017)Jia, Raphenya, Alcock, Waglechner, Guo, Tsang, Lago, Dave, Pereira, Sharma, Doshi, Courtot, Lo, Williams, Frye, Elsayegh, Sardar, Westman, Pawlowski, Johnson, Brinkman, Wright, and McArthur}]{Jia2017CARD}
Baofeng Jia, Amogelang~R. Raphenya, Brian Alcock, Nicholas Waglechner, Peiyao Guo, Kara~K. Tsang, Briony~A. Lago, Biren~M. Dave, Sheldon Pereira, Arjun~N. Sharma, Sachin Doshi, Mélanie Courtot, Raymond Lo, Laura~E. Williams, Jonathan~G. Frye, Tariq Elsayegh, Daim Sardar, Erin~L. Westman, Andrew~C. Pawlowski, Timothy~A. Johnson, Fiona~S.L. Brinkman, Gerard~D. Wright, and Andrew~G. McArthur. 2017.
\newblock \href {https://doi.org/10.1093/nar/gkw1004} {Card 2017: expansion and model-centric curation of the comprehensive antibiotic resistance database}.
\newblock \emph{Nucleic Acids Research}, 45(D1):D566--D573.

\bibitem[{Lakin et~al.(2019)Lakin, Kuhnle, Alipanahi, Noyes, Dean, Muggli, Raymond et~al.}]{lakin2019hierarchical}
Steven~M. Lakin, Alan Kuhnle, Bahar Alipanahi, Noelle~R. Noyes, Chris Dean, Martin Muggli, Rob Raymond, et~al. 2019.
\newblock \href {https://doi.org/10.1038/s42003-019-0545-9} {Hierarchical hidden markov models enable accurate and diverse detection of antimicrobial resistance sequences}.
\newblock \emph{Communications Biology}, 2(1):294.

\bibitem[{Lobo(2008)}]{lobo2008blast}
Ingrid Lobo. 2008.
\newblock Basic local alignment search tool (blast).
\newblock \emph{Nature Education}, 1(1):215.
\newblock © 2008 Nature Education.

\bibitem[{Luo et~al.(2022)Luo, Sun, Xia, Qin, Zhang, Poon, and Liu}]{luo2022biogpt}
Renqian Luo, Liai Sun, Yingce Xia, Tao Qin, Sheng Zhang, Hoifung Poon, and Tie-Yan Liu. 2022.
\newblock \href {https://doi.org/10.1093/bib/bbac409} {Biogpt: Generative pre-trained transformer for biomedical text generation and mining}.
\newblock \emph{Briefings in Bioinformatics}, 23(6):bbac409.

\bibitem[{Marini et~al.(2022)Marini, Oliva, Slizovskiy, Das, Noyes, Kahveci, Boucher, and Prosperi}]{marini2022amrmeta}
Simone Marini, Marco Oliva, Ilya~B Slizovskiy, Rishabh~A Das, Noelle~Robertson Noyes, Tamer Kahveci, Christina Boucher, and Mattia Prosperi. 2022.
\newblock \href {https://doi.org/10.1093/gigascience/giac029} {Amr-meta: A k -mer and metafeature approach to classify antimicrobial resistance from high-throughput short-read metagenomics data}.
\newblock \emph{GigaScience}, 11.
\newblock Giac029.

\bibitem[{Yoo et~al.(2024)Yoo, Sokhansanj, Brown, and Rosen}]{yoo2024predicting}
Hyunwoo Yoo, Bahrad Sokhansanj, James~R. Brown, and Gail Rosen. 2024.
\newblock Predicting anti-microbial resistance using large language models.
\newblock \emph{arXiv preprint arXiv:2401.00642}.
\newblock \url{https://doi.org/10.48550/arXiv.2401.00642}.

\bibitem[{Zhou et~al.(2023)Zhou, Ji, Li, Dutta, Davuluri, and Liu}]{zhou2023dnabert2}
Zhihan Zhou, Yanrong Ji, Weijian Li, Pratik Dutta, Ramana Davuluri, and Han Liu. 2023.
\newblock \href {https://arxiv.org/abs/2306.15006v1} {Dnabert-2: Efficient foundation model and benchmark for multi-species genome}.
\newblock \emph{arXiv}.
\newblock ArXiv:2306.15006v1.

\end{thebibliography}

\appendix

\section{Appendix}
\label{sec:appendix}

\subsection{Example Prompt Explanation including DNA Sequence}

In this example prompt, a DNA sequence is provided along with several drug class labels, such as Sulfonamides, Aminoglycosides, Betalactams, Glycopeptides, Tetracyclines, Phenicol, Fluoroquinolones, MLS (Macrolide-Lincosamide-Streptogramin), and Multi-drug resistance. The task involves asking the model to determine the drug class that the DNA sequence is resistant to. 

The prompt follows this format:

\begin{quote}
\texttt{"Tell me the resistance drug among drugs (Sulfonamides, Aminoglycosides, Betalactams, Glycopeptides, Tetracyclines, Phenicol, Fluoroquinolones, MLS, Multi-drug\_resistance) with DNA sequence (ATGAATCCCTATC...
...ACAAACTGCGAGGCAGTTCGCATGA)?"}
\end{quote}

This prompt is used to assess the DNA sequence for antibiotic resistance and classify the sequence into one of the specified drug resistance categories.

\subsection{Example Prompt Explanation including Blastn information}

In this prompt, a DNA sequence and the top 5 Blastn search results are provided. The task is to predict the drug class that the DNA sequence is resistant to, based on the alignment information and matching sequences. The drug class labels included in the prompt are Sulfonamides, Aminoglycosides, Betalactams, Glycopeptides, Tetracyclines, Phenicol, Fluoroquinolones, MLS (Macrolide-Lincosamide-Streptogramin), and Multi-drug resistance.

The BLASTn results contain gene information such as sequence titles, alignment length, e-values, and detailed sequence alignments (query, match, and subject sequences). This allows the model to analyze the DNA sequence's pattern and classify it into the appropriate drug resistance category.

The prompt follows this format:

\begin{quote}
\texttt{"Tell me the resistance drug among drugs (Sulfonamides, Aminoglycosides, Betalactams, Glycopeptides, Tetracyclines, Phenicol, Fluoroquinolones, MLS, Multi-drug\_resistance) with DNA information ([\{'sequence\_title': 'gi|1035502645|ref|NG\_048504.1| Enterococcus casseliflavus vanXY-C gene for D-Ala-D-Ala dipeptidase/D-Ala-D-Ala carboxypeptidase VanXY-C, complete CDS', 'alignment\_length': 673, 'e\_value': 0.0, 'query\_sequence': 'ATGAATCCCTATCTA...', 'match\_sequence': '||||||||||||||...', 'subject\_sequence': ...'\}, ... ])?"}
\end{quote}

This prompt aims to predict the antibiotic resistance drug by using DNA sequence data from the Blastn search results and identifying the relevant drug resistance class.

\end{document}